\pdfoutput=1
\documentclass[11pt]{article}

\usepackage{acl}

\usepackage{times}
\usepackage{latexsym}
\usepackage{amsmath}
\usepackage{multirow}

\usepackage{amssymb}

\usepackage[T1]{fontenc}
\usepackage[utf8]{inputenc}

\usepackage{microtype}

\usepackage{inconsolata}

\usepackage{graphicx}

\title{Multi-head Span-based Detector for AI-generated Fragments in Scientific Papers}

\author{German Gritsai\textsuperscript{1,2}
        \And 
        Ildar Khabutdinov\textsuperscript{1,2} \\ Advacheck OÜ, Estonia\textsuperscript{1} \\  Université Grenoble Alpes, France\textsuperscript{2} \\ \texttt{\{gritsai, khabutdinov, grabovoy\}@advacheck.com} \\ 
        \And
        Andrey Grabovoy\textsuperscript{1}}

\begin{document}
\maketitle
\begin{abstract}
This paper describes a system designed to distinguish between AI-generated and human-written scientific excerpts in the DAGPap24 competition hosted within the Fourth Workshop on Scientific Document Processing. In this competition the task is to find artificially generated token-level text fragments in documents of a scientific domain. Our work focuses on the use of a multi-task learning architecture with two heads. The application of this approach is justified by the specificity of the task, where class spans are continuous over several hundred characters. We considered different encoder variations to obtain a state vector for each token in the sequence, as well as a variation in splitting fragments into tokens to further feed into the input of a transform-based encoder. This approach allows us to achieve a 9\% quality improvement relative to the baseline solution score on the development set (from 0.86 to 0.95) using the average macro $F_{1}$-score, as well as a score of 0.96 on a closed test part of the dataset from the competition.
\end{abstract}

\section{Introduction}
Modern advances in the field of text generation models provide high quality artificial texts that are hardly distinguishable from human-written texts at fluent reading. Text generation systems such as Llama \cite{llama}, ChatGPT \cite{chat}, Mistral \cite{mistral} are increasing the list of successfully solved problems in updated benchmarks \cite{bench} with the release of each newer version of the product. Although the progress of such models is impressive, it poses new challenges for scientists, as the development of these systems implies the emergence and spread of generated fragments in texts of the scientific domain. In the field of Natural Language Processing there are already a large number of methodologies for detecting texts generated by machine learning models \cite{survey}, including approaches for the scientific domain \cite{sc_domain, gr_scien}. Therefore, the improvement of artificial text detection techniques occurs simultaneously with the improvement of text generation methods. In order to prevent the gap between quality generation and precision of detection from growing, it is necessary to periodically update and modernise existing detection approaches with newly generated fragments. Indeed, increasing the appearance of generated fragments in scientific papers carries with it a potential increase in plagiarism \cite{anypair}, fake studies \cite{fake} and missinformation.

The DAGPap24 competition allowed us to perform a further cycle of updates and make a reliable AI-generated scientific text detection system. The challenge is to build a detection system that is robust to generated fragments from fundamentally different frameworks and diverse scientific domains. 

In this paper, we present a solution that was developed for the DAGPap24 competition by our team. We propose a method to find artificially generated fragments at the token level using a multi-task architecture. Our model has two classifiers for each token, that are trained jointly and allow the model to have a higher generalisation capability and to process the text sequences arriving at the input faster. We introduce the aforementioned architecture in this paper and conduct experiments to tune its components to obtain the best result.

\section{Data and task description}
The DAGPap24 competition has been formulated as one classification task for the English language. Participants had to identify intervals in the scientific text that were human-written or machine-generated accordingly. The intervals are not fixed in advance and have to be labelled automatically by the detection algorithm based on the context. The problem has been proposed to be solved by token classification approaches from the competition organisers. A large number of works on artificial text detection is dealing with the classification problem at sequence level \cite{ippolito}. However, the approach of token classification and their subsequent grouping into intervals of the same class is also able to show a competitive performance in this task, in some cases producing better results than the established approach \cite{comarison}. According to the organisers the task contains scientific texts whose tokens should be classified into one of four classes: human, generated by ChatGPT, generated using synonyms from NLTK \cite{nltk}, generated using an unnamed summarisation model.
\subsection{Task Definition}
In this competition, it is expected that the problem of detecting machine-generated fragments will be solved as a token classification task. There is a given dataset \( \mathcal{D} = \{(x_i, y_i)\}_{i=1}^{N} \), where each document $x_i$ is represented by a finite combination of tokens:
\[
 x_i = \{x_i^1, \ldots , x_i^m \}, \scalebox{0.45}{\qquad} x_i^j \in \mathcal{W}, \scalebox{0.45}{\qquad} j \in \{1, \ldots , m\},
\]
\[
 y_i = \{y_i^1, \ldots , y_i^m \}, \qquad y_i^j \in \{0, 1, 2, 3\},
\]
where \(\mathcal{W}\) corresponds to chosen tokens vocabulary. The labels \( y_i^j \in \{1, 2, 3\}\) correspond to tokens that are likely machine-generated with synonym-replacement, ChatGPT-generated or summarized respectively and \( y_i^j = 0\) corresponds to human-written ones.

Formally, the task is to find a multiclass classifier that minimizes the empirical risk on the dataset \(\mathcal{D}\):
\begin{align}
\hat{g} = \underset{\substack{g \in \mathfrak{F}}}{\text{argmin}} \sum_{i=1}^{N} \sum_{j}^{M_i} \scalebox{0.10}{\qquad} [g(x_i^j) \neq y_i^j],
\end{align}
where $M_i$ is a number of tokens in $i$-th document, \(\mathfrak{F}\) is a set of all considered classification models.

\subsection{Data}
The organisers of the competition provided a training dataset that contained 5000 samples. Each sample was represented by text, annotations, tokens of the corresponding text and labels for each token. Example rows from the dataset can be seen in Table \ref{citation-guide}.

\begin{table*}
  \centering
  \begin{tabular}{p{4cm}p{4cm}p{3cm}p{3cm}}
    \hline
    \textbf{Text}           & \textbf{Annotations} & \textbf{Tokens} & \textbf{Token Label Ids} \\
    \hline
    The number of osteoporotic fractures ... & [[0, 3264, 'chatgpt'], ...] & ['The', 'number', 'of', 'osteoporotic', 'fractures', ...] & [2, 2, 2, 2, 2, 2,  ...] \\
    Blade surface roughness ranks amongst ... & [[0, 34694, 'human'], ...] & ['Blade', 'surface', 'roughness', 'ranks', 'amongst', 'the', ...] & [0, 0, 0, 0, 0, 0, ...] \\

    \hline
  \end{tabular}
  \caption{\label{citation-guide}
    A pair of examples of string representations in the provided dataset.  Each sample is represented by a text string, its markup, partitioning into tokens and labels to each token.
  }
\end{table*}

All texts in the rows of the dataset are documents from the scientific domain in which some parts have been replaced by machine-generated sequences. Fake text can even start in the middle of a sentence, so it is important to take the context into account when doing classification.

\begin{table}
  \centering
  \begin{tabular}{lp{1cm}p{1.4cm}p{1.38cm}}
    \hline
    \textbf{Label} & \textbf{Count} & \textbf{Mean length symbols} & \textbf{Mean length tokens}  \\
    \hline
    Human & 13346 & 10054.69 & 4370.62 \\
    NLTK-replace & 4245 & 3773.91 & 518.54 \\
    ChatGPT & 4447 & 2599.63 & 353.37 \\
    Summarized & 4376 & 1597.54 & 333.13 \\\hline
  \end{tabular}
  \caption{Statistics collected for each class of the provided dataset to train the models. It can be seen that half of the texts are human-written.}
  \label{tab:accents}
\end{table}

Before starting to build a model solution to the problem, we analysed the provided dataset. Some of its statistics can be seen in Table \ref{tab:accents}. The texts written by humans are equalled by the remaining three classes in terms of quantity, so to balance the classes all the artificially generated samples could be combined into one group. The human-written texts remain, understandably, longer in terms of average length for both tokens and characters. What is also notable is the length and quality of the generated texts under the summarised label. The excerpts of this class are shorter than the others and when analysed visually they immediately catch the eye due to the lack of postprocessing of the summarization model, the texts are contaminated with unnecessary symbols.

\section{Approach}

\subsection{Model}

\begin{figure}[t]
  \centering
  \includegraphics[width=0.43\textwidth]{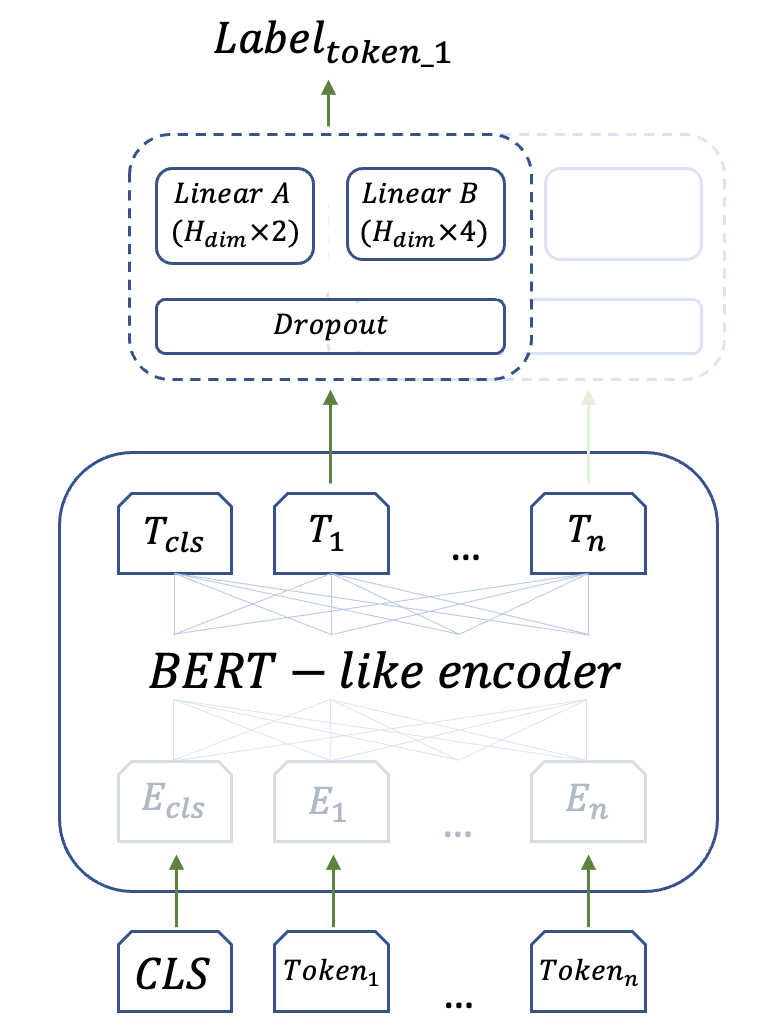}
  \caption{A multi-task architecture to solve the token classification problem. Each token, after receiving a vector for text representation via encoder, is classified in two heads: Linear A - binary and Linear B - multiclass.}
  \label{fig:model}
\end{figure}

There are several ways to solve the given problem, we have chosen the token classification approach. One of its subfields is Grammatical Error Correction (GEC) \cite{gec}, which aims to correct as many errors as possible in a given passage. It correlates with our task, where, for example, the junction from one class to another can be in an uncommon place and of any length, so we need to identify the boundaries very precisely. One state-of-the-art solution is the approach described by \cite{gector}. Their GECToR model has a custom multi-head transformer-based architecture with and solves the GEC problem with sequence tagging. We decided to use a similar architecture with our extension in the machine-generated fragments detection domain where we need to solve the problem at the token level. The architecture we use is shown in Figure \ref{fig:model}.

\subsection{Training and inference stages}
The token sequence that comes to the input of the algorithm is vectorised using a BERT-like encoder \cite{bert}, further the vector for each token is fed in parallel to the input of two linear layers with dropout. The major difference from the standard BERT-like approach for this task is the presence of two classification heads and multi-task learning. In our approach, one head (Linear A) is responsible for binary classification - human or machine-generated, while the second head (Linear B) solves the multi-class classification for the 4 classes that have been posed in the current competition. Thus, when training such a network,  the loss function aggregates the error values from both heads and sums them up.

Previous studies have captured that the multi-task learning approach reduces the probability of overfitting and increases the generalisability of the trained system \cite{height}. As for inference, we calculate the probability for each token of a given sequence to contain machine-generated data using \textit{Linear A} and softmax. If the maximum probability among all tokens of the sequence has a value that is higher than a predefined threshold, then classification of each token of the given sequence is performed by multiclass layer - \textit{Linear B}, otherwise we produce a label of human class. This approach on inference reflects the specificity of the current task, where, based on the statistics obtained in the previous section, it can be seen that texts are not interrupted by single insertions of a token from another class, and most often spans have long continuity intervals. Among the classes present, the length of human texts is most often higher than the others. Due to the limited number of characters that can be included in one sequence and fed to the BERT-like encoder input, sequences in this paradigm will most often contain only tokens of the same class. 

\section{Experiments}
The organisers of the competition presented a basic solution: DistilBERT \cite{distill} and SciBERT \cite{scibert}. Both models were fine-tuned with the classical architecture and pipeline of the token classification task. Considering the limited time of the competition, we planned experiments with the architecture described in the previous section: varying the input sequence length, changing BERT-like encoders and selecting hyperparameters.

\subsection{Varying the input sequence length}
BERT-like models have a constraint on the length of the input sequence. In earlier studies in the machine-generated fragment detection task, we analysed the effect of input length on detection quality \cite{we}. In the current task, each example was accompanied by a sentence partitioning into word tokens. However, depending on the choice of tokeniser, these token-words can be broken down into different numbers of smaller tokens to feed into the model input. For example, the dictionaries of the tokeniser adapted for this domain - SciBERT - have a large number of complete token-words, which means that splitting the original ones into smaller units will be minimal. Given this fact, we need to make sure that after tokenisation by the chosen tokeniser, the sequence that comes to the input of the model is not clipped and does not lose information that is useful for training algorithms. For different models we tried different ways of partitioning. 

\subsection{Changing BERT-like encoders}
The baseline result that was shown with the SciBERT model displayed high score. Currently, there are a large number of pre-trained encoders that are able to show good basic quality solution to the problem. However, further pre-training under the specific domain performs differently for all models. This is the motivation behind the encoder variation approach. Looking back at the experience of \cite{gector}, who proposed the GECToR architecture, we also examined XLNet \cite{xlnet}. In addition, we tried the approach with the QLoRa adapter \cite{lora} pre-training for the token classification task with the large language model (LLM) Mistral. This language model, according to benchmark results, is capable of solving an impressive number of tasks at a high level. Domain-specific LLM adaptor training may sometimes circumvent the quality of problem solving with established encoder models, such as RoBERTa \cite{lroberta}, DeBERTa \cite{deberta}, etc.

\section{Results}

\begin{table*}
  \centering
  \begin{tabular}{lccc}
    \hline
    \multirow{2}{*}{BERT-like encoder + tokens partition length} & \multicolumn{3}{c}{\textbf{Development Set}} \\
     & P & R & $F_{1}$ \\
    \hline
    DistilBERT + 512 (baseline) & 0.86 & 0.83 & 0.84  \\
    SciBERT + 512 (baseline) & 0.89 & 0.86 & 0.87  \\
    \hline
    Mistral w. QLoRA + 512 & 0.93 & 0.89 & 0.91 \\
    XLNet + 512 & 0.91 & 0.92 & 0.91  \\ 
    XLNet + 350 & 0.96 & 0.95 & 0.95  \\ 
    SciBERT + 185 & 0.92 & 0.89 & 0.91  \\ 
    SciBERT + 300 & 0.94 & 0.93 & 0.93  \\ 
    SciBERT + 350 & \textbf{0.97} & \textbf{0.96} & \textbf{0.96}  \\ 
    SciBERT + 400 & 0.93 & \textbf{0.96} & 0.94  \\ 
    SciBERT + 512 & 0.89 & 0.93 & 0.9  \\ 
    \hline
  \end{tabular}
  \caption{Results on a development dataset with different settings for the proposed multi-task architecture.}
  \label{tab:result}
\end{table*}

The results of the experiments can be seen in Table \ref{tab:result}. The specificity of the task, scientific domain, allowed us to keep the SciBERT model as the main encoder for our architecture after the experiments. However, as shown in the table, the best quality is achieved when the texts are divided into intervals of 350 token-words. Given the limited input context of transformed-based models, it is the case that the least amount of information is lost after tokenisation, thus achieving the highest quality. In addition, the dropout value was selected, and after varying it was set to 0.7. Also it was observed that with the threshold value set to 0.55, a quality improvement is obtained to drop human-written sequences during inference. The previously described architecture with two linear layers together with these settings helped our team to achieve the quality of 0.96 average macro $F_{1}$-score on the development set. As for the test set, the results were similar and we took 5th place in the competition with an average of 0.96.

\section{Conclusion}

In this paper we present a descriptive overview of the solution approach that was used by our team to solve the DAGPap24 competition for detecting  AI-generated fragments in scientific documents. A token classification approach was chosen to map each token to one of four pre-defined classes. In our solution, we used a BERT-based encoder and two linear layers to process the output vectors of the encoder for each token in sequence. One of the layers is responsible for binary classification and the other for the predefined 4 classes. Exploring shared training, the classification quality becomes better for identical model settings, and on the provided development dataset, this method helped to improve the quality by 9\% on the macro $F_{1}$-score metric. 

In the future work we would be interested in analysing the distribution of the metric across classes and based on this we would like to add weights to the loss function during pre-training for class balancing. In addition, it would be interesting to observe this approach on a larger dataset, where classes can be defined by more common text generation methods.

\bibliography{main}

\end{document}